\documentclass[10pt,twocolumn,letterpaper]{article}
\usepackage[accsupp]{axessibility}
\usepackage[pagenumbers]{wacv} 

\usepackage{graphicx}
\usepackage{amsmath}
\usepackage{amssymb}
\usepackage{booktabs}
\usepackage{enumitem}
\usepackage{chapterbib}
\usepackage{colortbl}
\usepackage{color}
\usepackage[dvipsnames]{xcolor}
\definecolor{Gray}{gray}{0.77}
\definecolor{DarkGray}{gray}{0.43}
\definecolor{LLGray}{gray}{0.91}
\definecolor{LGray}{gray}{0.94}
\definecolor{LightCyan}{rgb}{0.88,1,1}
\definecolor{Greenish}{rgb}{0.10,0.60,0.30}
\definecolor{ForestGreen}{RGB}{34,139,34}
\definecolor{Wood}{RGB}{150,111,51}
\definecolor{DarkBrown}{RGB}{150,78,2}
\definecolor{Pinkish}{RGB}{255,102,220}
\definecolor{Blueish}{RGB}{2,78,150}

\newcolumntype{a}{>{\columncolor{Gray}}c}
\newcolumntype{g}{>{\columncolor{LGray}}c}

\usepackage[pagebackref,breaklinks,colorlinks]{hyperref}

\usepackage[capitalize]{cleveref}
\crefname{section}{Sec.}{Secs.}
\Crefname{section}{Section}{Sections}
\Crefname{table}{Table}{Tables}
\crefname{table}{Tab.}{Tabs.}

\def\wacvPaperID{86} 
\def\confName{WACV}
\def\confYear{2024}

\begin{document}


\title{Investigating the Role of Attribute Context in Vision-Language Models for Object Recognition and Detection}


\author{    Kyle Buettner\textsuperscript{\rm 1},
    Adriana Kovashka\textsuperscript{\rm 1,2}\\
\textsuperscript{1}Intelligent Systems Program, \textsuperscript{2}Department of Computer Science,
University of Pittsburgh, PA, USA\\
{\tt\small buettnerk@pitt.edu, kovashka@cs.pitt.edu}
}
\maketitle
%



\begin{abstract}

   Vision-language alignment learned from image-caption pairs has been shown to benefit tasks like object recognition and detection. Methods are mostly evaluated in terms of how well object class names are learned, but captions also contain rich attribute context that should be considered when learning object alignment. It is unclear how methods use this context in learning, as well as whether models succeed when tasks require attribute and object understanding. To address this gap, we conduct extensive analysis of the role of attributes in vision-language models. We specifically measure model sensitivity to the presence and meaning of attribute context, gauging influence on object embeddings through unsupervised phrase grounding and classification via description methods. We further evaluate the utility of attribute context in training for open-vocabulary object detection, fine-grained text-region retrieval, and attribution tasks. Our results show that attribute context can be wasted when learning alignment for detection, attribute meaning is not adequately considered in embeddings, and describing classes by only their attributes is ineffective. A viable strategy that we find to increase benefits from attributes is contrastive training with adjective-based negative captions. 
   
\end{abstract}

\section{Introduction}
    \label{sec:intro}

    \begin{figure}[t]
    \centering
    \includegraphics[scale=0.305]
    {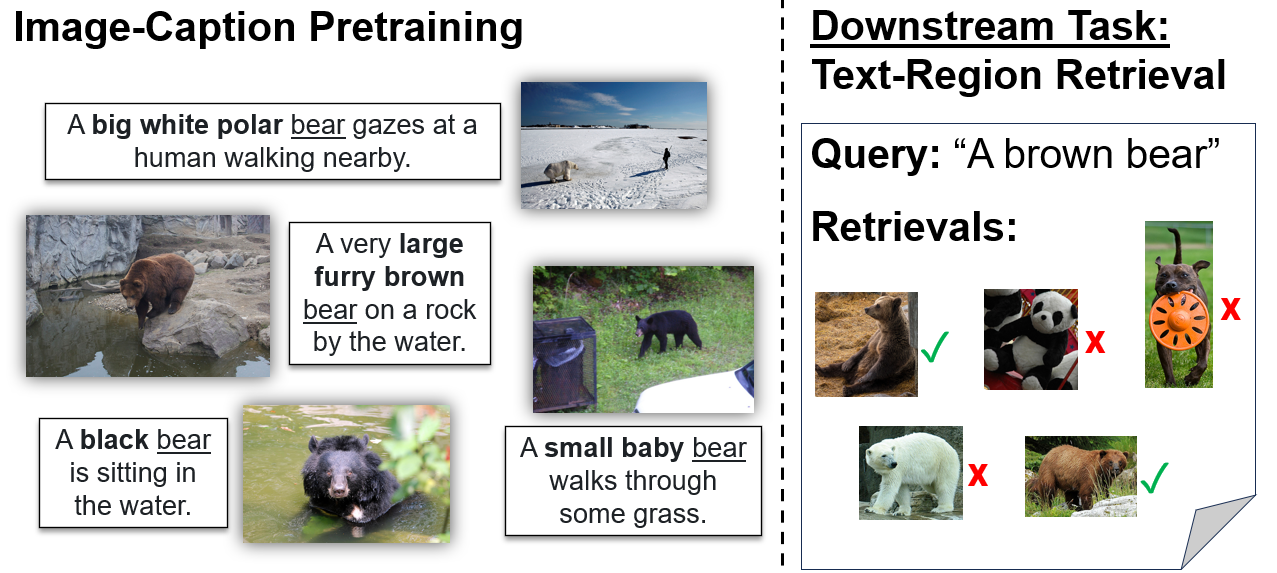}
    \vspace{-2mm}
    \caption{\textbf{Do vision-language models effectively leverage attribute context in captions?} In captions, objects (\eg \underline{bear}) are often described with rich contextual information (\eg attributes like \textbf{big}, \textbf{furry}, \textbf{black}). We evaluate the impact and utility of attributes in VL modeling through tasks such as text-region retrieval. }
    \label{intro_fig}
\end{figure}

    Natural language has been shown to provide a strong signal for training visual representations. A visual-text alignment model can be pretrained with image-caption data and used for downstream tasks like object recognition, detection, and retrieval. While the text embeddings that represent object nouns are often used as classifier weights, the impact and utility of other caption context, especially attributes, are less clear. Consider the Fig. \ref{intro_fig} caption: ``A very \textbf{large} \textbf{furry} \textbf{brown} \underline{bear} on a \underline{rock} by the \underline{water}.''
    The model can learn grounding using only nouns (underlined), but \textit{bear} can also be learned in the context of its attributes (bolded adjectives). Do alignment models use attribute context to learn \textit{bear}? Do they distinguish between \textit{brown bear} and \textit{black bear}?

    Attributes in captions can aid object recognition and detection in various ways. Attributes can serve as a proxy for a fine-grained category which is not explicitly mentioned (\eg a \textit{small, young cat} is a \textit{kitten}). They can ensure that the alignment model is paying attention to the right features, rather than dataset artifacts (\eg that a \textit{bear} is being grounded as such because it is \textit{brown/black}, rather than because its background is a \textit{forest}). They can be used to specify subcategories, \eg when a user desires detections that match a certain property (\eg a \textit{red car}, but not a \textit{blue car}).  

    With this motivation, our goal is to better understand the connection between attributes and objects in vision-language (VL) models. In particular, we explore considerations such as whether models leverage attributes in captions as important signals for object learning and whether VL models can use object and attributes effectively in fine-grained tasks (\eg recognizing ``a large bird with a brightly colored bill"). We refer to a model's capabilities in using attribute information as \textit{attribute sensitivity}.
    
    We offer an extensive sensitivity analysis of two popular alignment paradigms for VL models, region-word grounding and whole image-text alignment, which we study with OVR-CNN \cite{zareian2021open} and CLIP \cite{radford2021learning}, respectively. We investigate if model decisions consider attribute presence and meaning, specifically testing how attribute perturbations affect unsupervised phrase grounding and classification via description tasks. \textit{We find an overall lack of sensitivity to attribute meaning}, inspiring an investigation into adjective-based contrastive negative sampling of captions. Our exploration results in strategies that increase the benefits from attribute context, exhibited through improvements in practical use cases of open-vocabulary object detection, fine-grained text-region retrieval, and object attribution.

    Our main contributions are insights into these questions:
    \begin{enumerate}[nolistsep,noitemsep]

         \item Does attribute context play an impactful role in VL pretraining for object detection? 

         \item Does learning to ground objects to contextualized word embeddings utilize attribute meaning? 

         \item Do VL models perform well at tasks where objects are described with/in terms of attributes? 

         \item Can contrastive negative sampling of captions increase a model's ability to use attribute context? 
         
         \item What sampling mechanisms are most effective? 

    \end{enumerate}

\section{Background and related work}
\label{sec:relatedwork}

   \noindent \textbf{Visual representation learning with language} Common VL tasks such as phrase grounding and visual question answering leverage objectives that align and/or merge image and text features \cite{alayrac2022flamingo, kim2021vilt, li2022blip, li2021align, wang2022image, wang2021simvlm,  yu2022coca, yuan2021florence}. Alignment achieved with large-scale contrastive learning \cite{hadsell2006dimensionality} has powered the traditional vision task of image classification by enabling impressive zero-shot capability (\eg ALIGN \cite{jia2021scaling}, CLIP \cite{radford2021learning}). The primary mechanism for adapting models like CLIP to recognition is through creating prompts (\eg ``a photo of a [classname]")  for all classes and using the text encoder to convert prompts into classifier weights. Recent methods have exploited this open-vocabulary capability of CLIP to provide attribute context with LLM-based class descriptions \cite{menon2022visual, pratt2022does}. It is still unclear the extent to which VL models for recognition can consider attributes. As such, we conduct more in-depth experiments within the ``classification via description" task of \cite{menon2022visual}, highlighting limited utility of attributes in zero-shot recognition with CLIP. 
      
   Our analysis also hones in on object detection, which typically entails a predefined class list and bounding box annotations. The use of text embeddings has expanded the detection vocabulary \cite{kamath2021mdetr, li2022grounded, zhang2022glipv2, bansal2018zero}, and large-scale image-caption datasets and weakly supervised objectives have enabled ``cheaper" supervision  \cite{ye2019cap2det,desai2021virtex}. Open-vocabulary detection \cite{zareian2021open}, which involves training on base classes and using region-text alignment to extend to novel classes, has become especially popular. Recent open-vocabulary detectors leverage CLIP through mechanisms such as distillation, prompting, and pseudo-labeling \cite{bangalath2022bridging, du2022learning, gao2022open, gu2021open, wu2023aligning, wu2023cora, zhong2022regionclip, zhou2022detecting}. Our work impacts this area as we gauge the attribute sensitivity of the CLIP model widely used with these approaches. Additionally, through \cite{zareian2021open}, we study fine-grained region-word alignment \cite{chen2020uniter, kim2021vilt, li2022grounded}, which is tailored to region-level tasks \cite{yao2023detclipv2}. We in particular examine the impact of attribute context in detection through comparing results to \cite{zareian2021open}. 
   
    \noindent \textbf{Bias and sensitivity measurement of embeddings in VL tasks} Our work relates to efforts to understand the biases in embeddings from VL models. Such probing has highlighted that grounded/aligned embeddings encode social biases \cite{ross2020measuring} and lack sensitivity to composition and word order \cite{thrush2022winoground, yuksekgonul2022and}. Our investigation more thoroughly analyzes embeddings with respect to attributes. For example, we gauge whether visual embeddings are sensitive to attribute presence and meaning when grounding to contextualized word embeddings (from \cite{devlin2018bert}). The work of \cite{bertasius2020cobe} relates as it involves use of contextualized object embeddings to detect object states (\eg \textit{sliced tomato}, \textit{tomato in a bowl}). Our work instead explores if \textit{enhancing the attribute sensitivity} of contextualized object embeddings impacts more general object detection and fine-grained text-region retrieval tasks. 

    \noindent \textbf{Contrastive negative sampling} ``Hard" negative samples can benefit contrastive learning \cite{kalantidis2020hard, robinson2020contrastive}. We explore negative sampling of captions to enhance attribute context in region-word pretraining and CLIP finetuning. Past works have created negatives by replacing nouns \cite{gupta2020contrastive} and by perturbing word order \cite{yuksekgonul2022and}. We alternatively test more attribute-tailored strategies by replacing only adjectives in captions, randomly/plausibly based on a dataset. We exhibit that order perturbations \cite{yuksekgonul2022and} do \textit{not} help fine-grained text-region retrieval, while adjective negatives do. We also show that adjective negatives improve vs. a generic caption sampling baseline on Visual Genome Attribution \cite{yuksekgonul2022and},
    and that adjective negatives benefit detection in region-word pretraining. Concurrent work \cite{doveh2023teaching} has also shown the value of adjective negatives, though our work uniquely shows value in detection. \cite{cascante2023going} also leverages attribute perturbations, but alternatively with synthetic visual data. 

    \noindent \textbf{Attributes in vision tasks} Our work considers object attributes, described through adjectives in captions, with respect to object learning and model capabilities. Past work has explored attributes with respect to direct prediction \cite{pham2021learning}, compositional zero-shot recognition with objects \cite{saini2022disentangling, nayak2022learning}, and use as a bridge between base and novel classes in zero-shot classification \cite{lampert2013attribute, xu2020attribute}. With respect to localization, attributes have served as signals to spatially constrain object learning \cite{jerbi2020learning, xiao2017weakly} and as part of an open-vocabulary attribute detection task (detecting all attributes with an object) \cite{bravo2022open}. Our work is unique as we analyze attributes \textit{as context for objects}, gauging impact in tasks like retrieval and detection.

\section{Methodology}
    \label{sec:methods}

    When VL models learn alignment for recognition and detection, the \textit{utility} of attributes in captions is considerably underlooked. We study object representation sensitivity to attributes through case studies in region-word grounding (OVR-CNN \cite{zareian2021open}) and image-text alignment (CLIP \cite{radford2021learning}). This section outlines these frameworks, our measurement methods, and our strategies to enhance attribute context. 
    
\subsection{Vision-language frameworks of study}

    We study contrastive frameworks that learn in each iteration using a batch $\mathcal{B}$ of image-caption pairs ($\mathcal{B}_I$ and $\mathcal{B}_C$ for just images and captions, respectively). A score $\langle I, C\rangle$ is computed to quantify the relative matching between an image $I$ and a caption $C$. Image-to-text and text-to-image contrastive objectives are as shown in Eqs. \ref{gen_loss_i2t} and \ref{gen_loss_t2i}:

    \begin{equation} \label{gen_loss_i2t}
        { \mathcal{L}_{I \rightarrow T}(I) = - \log \frac{\exp \langle I, C\rangle}{\sum_{C'\in \mathcal{B}_C } \exp \langle I, C'\rangle} } 
    \end{equation}

    \begin{equation} \label{gen_loss_t2i}
        { \mathcal{L}_{T \rightarrow I}(C) = - \log \frac{\exp \langle I, C\rangle}{\sum_{I'\in \mathcal{B}_I } \exp \langle I', C\rangle} } 
    \end{equation}

    Methods may differ in terms of how the scoring function is defined and whether losses include additional components such as temperature or normalization constants.

    \subsubsection{Case study: Region-word grounding}
    \label{rwgroundcase}
        We explore region-word grounding to learn fine-grained alignment for detection. We specifically consider OVR-CNN \cite{zareian2021open}, which employs a weakly supervised, region-word grounding pretraining task to learn class embeddings for open-vocabulary detection with Faster R-CNN \cite{ren2015faster}. The model resembles PixelBERT \cite{huang2020pixel}, using ResNet-50 \cite{he2016deep}, a pretrained BERT \cite{devlin2018bert}, and a BERT-like multimodal model. The total loss $\mathcal{L}(I, C)$ comprises four objectives: masked language modeling ($\mathcal{L}_{MLM}$), image-to-text matching ($\mathcal{L}_{ITM}$), and two contrastive grounding terms ($\mathcal{L}_{G}(C)$ and $\mathcal{L}_{G}(I)$). $\langle I,C \rangle$ for OVR-CNN is defined with Eqs. \ref{groundingscore_contextfree} and \ref{activitycoeff}, where the dot product is taken between each word token $e_{j}^{C}$ ($n_C$ total produced from BERT's input layer) and each region token $e_{i}^{I}$ ($n_I$ total from ResNet then projected into the language embedding space with a V2L layer):
    
        \begin{equation}        
            \label{groundingscore_contextfree}
                {\langle I,C \rangle  = \frac{1}{n_C}\sum_{j=1}^{n_C}  \sum_{i=1}^{n_I} a_{i,j} \langle e_{i}^{I}, e_{j}^{C} \rangle} 
        \end{equation}
    
        \begin{equation}        
            \label{activitycoeff}
                { a_{i,j} = \frac{\exp \langle e_{i}^{I}, e_{j}^{C} \rangle}{\sum_{i'=1}^{n_I} \exp \langle e_{i'}^{I}, e_{j}^{C} \rangle} } 
        \end{equation}    
    
        Notably, this default alignment mechanism uses \textit{context-free} word embeddings ($e_{j}^{C}$ in BERT), which do not change with surrounding language context (\eg \textit{orange} has the same embedding in the captions ``\underline{orange} basketball" and ``eating an \underline{orange}"). We reason that this type of grounding contributes to misalignment of concepts, potentially inhibiting the benefits of attribute context. More recent models (\eg CLIP \cite{radford2021learning}) also align visual regions to text embeddings contextualized through transformers. For expansive insights, we experiment with contextualization in OVR-CNN by altering Eq.~\ref{groundingscore_contextfree} to use $f_{j}^{C}$, which are BERT's \textit{output} embeddings that change with context (unlike BERT's $e_{j}^{C}$ which are static). Eq. \ref{groundingscore_contextualized} shows this change:

        \begin{equation} \label{groundingscore_contextualized}
        	{\langle I,C \rangle = \frac{1}{n_C}\sum_{j=1}^{n_C}  \sum_{i=1}^{n_I} a_{i,j} \langle f_{i}^{I}, f_{j}^{C} \rangle } 
    	\end{equation}
    
        Since word embeddings are dynamically contextualized, visual regions for an object are grounded to a \textit{collection} of embeddings instead of one. Naive integration of such embeddings into detection results in poor performance. We use the following training recipe to effectively use contextualized embeddings in detection: (1) using a prompt ``A/an $<$objName$>$." when changing a class embedding for object $k$ from $e_{k}^{\mathcal{C}}$ to $f_{k}^{\mathcal{C}}$, (2) allowing the language encoder to update in the grounding pretraining task, and (3) allowing the V2L layer to update in finetuning. These strategies provide the training flexibility needed to thoroughly evaluate attribute sensitivity with contextualized embeddings.

    \subsubsection{Case study: CLIP image-text alignment}
    
         Open-vocabulary detectors that have come after OVR-CNN notably leverage CLIP \cite{bangalath2022bridging, du2022learning, gao2022open, gu2021open, wu2023aligning, wu2023cora, zhong2022regionclip, zhou2022detecting}. Their ability to use attribute context is thus highly dependent on the attribute sensitivity of CLIP. We study CLIP's attribute sensitivity for insights that generalize to various methods. The alignment objective of CLIP notably differs from OVR-CNN in that it aligns embeddings corresponding to entire images and text descriptions rather than to regions and words. More specifically, an image $I$ and caption $C$ are processed by CLIP's image and text encoders to produce normalized feature representations $z_i^I$ and $z_j^C$. $\langle I,C \rangle$ for CLIP is defined in Eq. \ref{groundingscore_clip}, where $\langle z_i^I,z_j^C\rangle$ is a dot product:

        \begin{equation}        
            \label{groundingscore_clip}
        	{\langle I,C \rangle  = \langle z_i^I,z_j^C \rangle } 
	  \end{equation}

        A temperature $\tau$ is also used with the losses in Eqs. \ref{gen_loss_i2t} and \ref{gen_loss_t2i}. Due to CLIP's large size and scale, we focus on finetuning representations, rather than pretraining from scratch. 
        
\subsection{Analyzing model sensitivity to attributes}
    \label{sens_method}

\begin{figure}[t]
    \centering
    \includegraphics[scale=0.24]{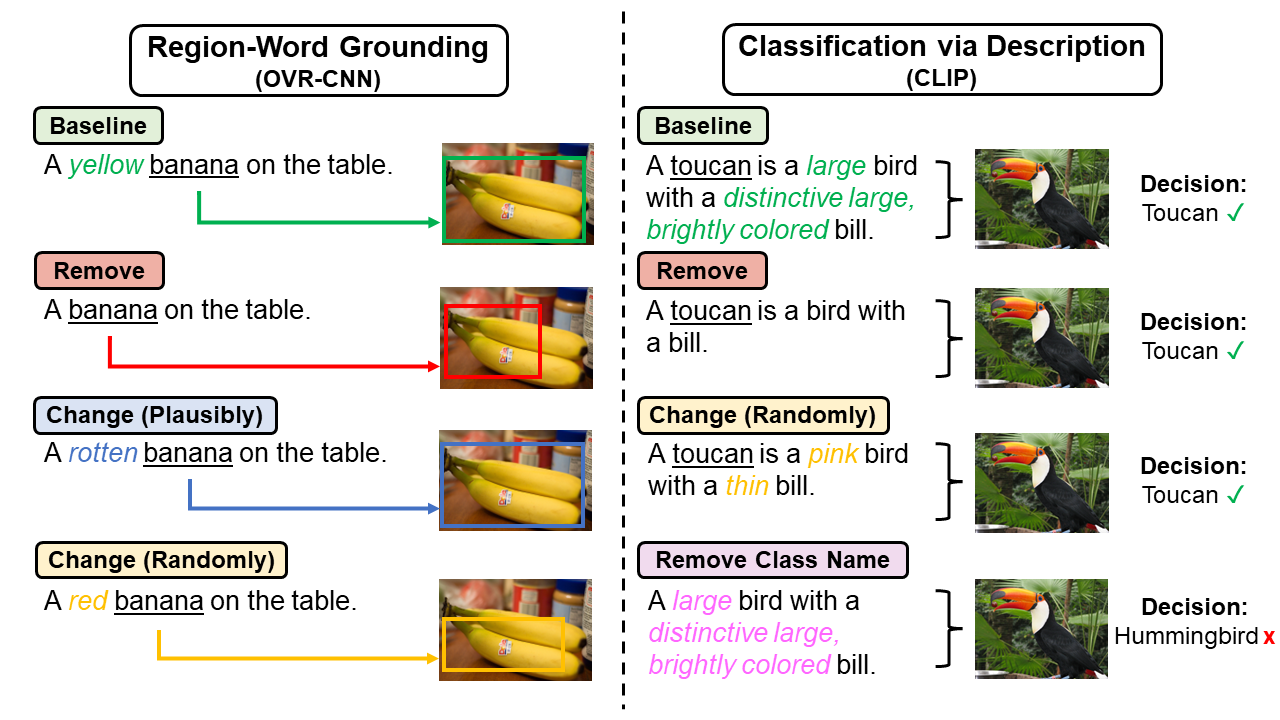}
    \vspace{-3mm}
    \caption{\textbf{Our attribute sensitivity measurement methodology.} We remove/change attributes/class names in text for grounding and classification tasks to measure if model decisions are sensitive to attributes. We show example predictions when attribute meaning is ignored (\eg \textit{rotten} \underline {banana}, \underline{toucan} is a \textit{pink} bird). }
    \label{attr_sens}
\end{figure}

      We aim to measure how influential attribute context is to a model's decision (\eg classification, grounding). We reason that in an attribute-sensitive model, the \textit{presence} of attributes should help decisions, as this information is complementary to objects. Additionally, the \textit{meaning} of attributes should be respected. Object representations should be more aligned when correct attributes are used than when incorrect attributes are used. Our mechanism for exploring these considerations is through \textit{removing} and \textit{changing} attribute context in the text for a task, as removal tests presence, and changing tests meaning. In this section, we outline our measurement methodology, for which we explore prior tasks that can show attribute sensitivity while fitting each alignment mechanism. In particular, we use unsupervised phrase grounding \cite{nebbia2022doubling} for OVR-CNN and classification via description \cite{menon2022visual} for CLIP, as shown in Fig. \ref{attr_sens}. 

     \noindent \textbf{Isolating objects and attribute context} For analysis of OVR-CNN (and for training as outlined in Sec. \ref{enhance}), we define a vocabulary $\mathcal{V}$ to be the nouns corresponding to objects in a dataset $\mathcal{D}$. In our study, $\mathcal{D}$ is COCO \cite{chen2015microsoft}, with 118,287 images and 5 captions per image. We build $\mathcal{V}$ from the synonym list of COCO class names provided in \cite{lu2018neural}, with plural terms added. The vocabulary $\mathcal{V}$ captures various terms for each class (\eg \textit{jet}, \textit{aircraft}, \textit{planes} for \textit{airplane}). We identify a class attribute as any adjectival modifier (``amod") with dependency on a class synonym in $\mathcal{D}$, detected with \cite{Honnibal_spaCy_Industrial-strength_Natural_2020}. The unique adjectives for each class make up respective \textbf{plausible} sets, containing attribute properties across the dataset (\eg a \textit{frisbee} is \textit{red}/\textit{green}/etc.). Unique adjectives across all classes make up the \textbf{random} set. We provide further details and statistics in the supp. material.
          
     \noindent \textbf{Measuring attribute sensitivity in region-word grounding} OVR-CNN is analyzed using unsupervised phrase grounding \cite{nebbia2022doubling}, a task that returns a bounding box $b$ for a text query $t$. Given an image-caption pair ($I$, $C$), we ask: if $I$ has a \textit{red car}, are visual regions for that car grounded better when using the car embedding in the caption ``a red \underline{car}..." than when using the embedding in ``a blue \underline{car}..." or ``a \underline{car}..."? Put another way, we test if the model leverages attribute meaning when grounding object regions to contextualized word embeddings. While a model could align visual regions for \textit{car} independently of attributes (\eg with context-free embeddings), we reason that bag-of-words behavior may result since embeddings are the same in cases like ``a red car and blue truck"/``a blue car and red truck". Also, the model would not be fully leveraging capabilities of contextualized embeddings, where a region-word objective can encode attribute information within a contextualized object grounding, such that the model dynamically learns to represent \textit{red car} vs. \textit{blue car}. 

     In this setup, we test four grounding scenarios: (1) using the \textit{baseline caption}, containing ground-truth attributes in adjective form (\eg ``a \textit{yellow} \underline{banana} on the table"); (2) using a caption that has object adjectives \textit{removed} (\eg ``a \underline{banana} on the table"); (3) using a caption that has object adjectives changed \textit{plausibly} according to our sets (\eg ``a \textit{rotten} \underline{banana} on the table"); and (4) using a caption that has adjectives changed \textit{randomly} to be any intra-corpus (\eg ``a \textit{red} \underline{banana} on the table"). In an attribute-sensitive model, we expect the top-performing grounding to have the most information (\eg ``yellow \underline{banana}"). We expect removal performance to drop vs. this baseline as objects are less specified. We reason that changing adjectives should make attributes \textit{incorrect} and thus hurt vs. the baseline.
     In the plausible case, we expect the dataset to cover disjoint states (\eg \textit{wooden} vs. \textit{plastic} spoon). While multiple attributes could be valid for an object, in practice, we find such cases rare.
     On 100 random samples, we find that 84\% of captions changed plausibly and 92\% changed randomly are not reasonably correct. We expect changing plausibly to thus result in a smaller drop from the baseline vs. changing randomly.

     To compute groundings, for each caption token $j$, if it matches an object term in $\mathcal{V}$, one or more bounding boxes are generated from the binary map of region-word similarity $\langle f_{i}^{I}, f_{j}^{C} \rangle$  such that $\langle f_{i}^{I}, f_{j}^{C} \rangle$  $\geq th_{sim}$. In the supp. material, we test at values $th_{sim}$=5, 10, 15 and show trends are not sensitive to this threshold. Then for all captions which mention that object, these boxes are compared to the ground-truth at various IoU thresholds, producing AP@$t$ values. We use $t$=30,40,50 as non-aggressive thresholds suitable for unsupervised inference. The average AP@IoU=30:10:50 is reported over all classes in the COCO validation set.

     
    \noindent \textbf{Measuring attribute sensitivity in CLIP image-text alignment} We analyze CLIP's attribute sensitivity through classification via description\cite{menon2022visual}, which adds attribute context to object prompts to aid zero-shot inference. In \cite{menon2022visual}, for each class \textit{c} in a dataset $\mathcal{D}$, GPT-3 is prompted to produce a list of descriptors $D(c)$. The descriptors contain attributes relevant to the object, along with $c$ to condition the attributes. For instance, the descriptors produced for \textit{toucan} are ``a/an \underline{toucan} which (is/has/etc) \textit{large}, \textit{brightly colored} bill.", ``a/an \underline{toucan} which (is/has/etc) \textit{long}, \textit{pointed} wings.", etc. To classify an image $I$, each descriptor $d$ serves as a prompt. The score for each class is computed using the average CLIP logits, $\phi(I,d)$, over each $d$, shown in Eq. \ref{score_orig}:

    \begin{equation}
        \label{score_orig}
        s(c, I) = \frac{1}{|D(\textit{c})|}\sum_{d \in D(\textit{c})} \phi(I,d)
    \end{equation}
    
     We select $\mathcal{D}$ to be ImageNetV2 \cite{recht2019imagenet} and use GPT-3 (\textit{davinci-002}) to produce descriptors. We also test producing only a single-sentence description to simulate the related method \cite{pratt2022does} (\eg ``A \underline{toucan} is a \textit{large} bird with a \textit{distinctive large, brightly colored} bill."). With both setups, we test sensitivity through removing and changing, detected as ``ADJ" with \cite{Honnibal_spaCy_Industrial-strength_Natural_2020}, but unlike OVR-CNN, we only use random changing (and not plausible) since the descriptors do not have a vocabulary like COCO. Then to further stress test CLIP, in a given inference, we remove \textit{all} class names by replacing them with ``a/an object". This experiment gauges whether CLIP can interpret objects from attribute-only descriptions (\eg a \textit{small}, \textit{white}, \textit{round} object with \textit{red seams} is a \textit{baseball}). In the supp. material, we provide specific details, examples, and linguistic properties for descriptors.


\subsection{Enhancing model sensitivity to attributes}
    \label{enhance}

    We also hypothesize that \textit{enhancing} attribute context's role can help tasks like object detection and text-region retrieval. We specifically experiment with \textit{adjective-based negative caption sampling}, where a negative for a caption $C$ includes the same words, just with an adjective replaced (\eg for ``\textit{blue} car in the street", \textit{blue} is replaced with \textit{red}). We reason that these negatives can encourage models to capture attribute meaning when learning objects, increasing the model's fine-grained utility. In pretraining specifically, another benefit is that attributes may help ``guide" object grounding to the correct regions (\eg a \textit{car} to a \textit{red} region). 

    We test negative sampling in OVR-CNN pretraining and CLIP finetuning, both with COCO. We explore two replacement methods: (1) choosing a \textit{random} adjective from the corpus and (2) choosing a \textit{plausible} adjective for a noun, such that it is mentioned intra-dataset with the respective class term. Through these strategies, we aim to gauge whether it is beneficial to contrast disjoint states in a dataset with \textit{plausible} (\eg \textit{wooden} vs. \textit{metal spoon}) or if simple \textit{random} adjectives suffice.  Table \ref{example_caps} shows examples of plausible and random captions. To implement in training, for each caption in $\mathcal{B}_C$ with an adjective detected, a negative caption is added to a batch $\mathcal{B}_N$. The loss in Eq.~\ref{gen_loss_i2t} becomes:

    \begin{equation}        \label{groundingloss_image_with_negatives}
    	{ \mathcal{L}_{I \rightarrow T}(I) = - \log \frac{\exp \langle I, C\rangle}{\sum_{C'\in \mathcal{B_C} + \mathcal{B_N} } \exp \langle I, C'\rangle} } 
    \end{equation}

     A potential shortcut with region-word grounding is that a model can solve the task by grounding just adjectives rather than object words. To encourage OVR-CNN to consider objects and attributes, we use noun negatives (using the same caption, but replacing nouns with random ones from $\mathcal{D}$). For CLIP, if no adjective-noun pair is detected, we add a random caption to $\mathcal{B}_N$. We also compare the plausible and random strategies to order-perturbing sampling \cite{yuksekgonul2022and}, since order perturbations can influence attention to attributes (\eg ``red car and blue truck" vs. ``red truck and blue car"). We test this strategy by perturbing order when possible (i.e. the caption has adjectives and nouns); as with other strategies, we sample a random caption otherwise. 

     \setlength{\tabcolsep}{5pt}
 \renewcommand{\arraystretch}{0.94}
\begin{table}
    \begin{center}
    \begin{tabular}{c|c}

    \small Caption &    \small A bunch of {\color{ForestGreen}\textbf{green}}
    
    \underline{bananas} growing in a tree.
  \\
  \hline
   \small Plausible Neg. &    \small A bunch of {\color{DarkBrown}\textit{\textbf{rotten}}} \underline{bananas} growing in a tree.
  \\
    \hline
    \small Random Neg. &    \small A bunch of {\color{Pinkish}\textit{\textbf{pink}}} \underline{bananas} growing in a tree.
  \\
    \end{tabular}
    \end{center}
    \vspace{-6mm}
    \caption{Examples of negative adjective captions.}
    \label{example_caps}
\end{table}

 \section{Evaluation}

        We evaluate context enhancement on one object-focused task (open-vocabulary detection) and two fine-grained tasks (text-region retrieval/object attribution).

        \noindent \textbf{Datasets} For image-caption training, we use COCO Captions \cite{chen2015microsoft}, and for finetuning open-vocabulary detection, we use COCO Objects \cite{lin2014microsoft}, (2017 train/val for both).
        The class split for open-vocabulary detection is the same as \cite{bansal2018zero,zareian2021open} (48 base and 17 target classes). For retrieval, we use the 2,000 image COCO val subset with object and attribute annotations from OVAD \cite{bravo2022open}. For attribution, we use ARO Visual Genome Attribution (VGA) \cite{yuksekgonul2022and}, with 28,748 examples.

     \noindent \textbf{Open-vocabulary object detection} This task considers \textit{base}/\textit{target} class sets with/without bounding box annotations. A detector (i.e. Faster R-CNN \cite{ren2015faster}) is trained only on base classes, and there are three evaluation settings: 
     \textit{base} classes only, \textit{target} only, and \textit{generalized}. As in \cite{zareian2021open}, base only and target only classify over the respective set, while in generalized, prediction is performed over the union of base and target classes, and results are reported within each group and overall. We report $AP_{50}$ as the metric, as in \cite{zareian2021open}.

    \noindent \textbf{Text-region retrieval} We pose fine-grained text-to-region retrieval as a use case where attribute-object understanding is needed. We input a set of texts $\mathcal{T}$ for which each text \textit{t} contains an attribute \textit{a} from the set $\mathcal{A}$ and an object \textit{o} from the set  $\mathcal{O}$ (\eg \textit{red car}). The goal is to return as output top-scoring regions that are correct if they contain the correct attribute \textit{and} object (\eg for \textit{red car}, non-red cars would be incorrect). We select $\mathcal{A}$ to be colors, patterns (striped, dotted, etc.), and materials (metal, wooden, etc.) in OVAD \cite{bravo2022open} and $\mathcal{O}$ to contain all COCO objects. Since OVAD's annotations are dense, we exclude attribute-object pairs that are not described in language due to being inherent (\eg \textit{metal car}) and use attribute-object pairs with greater than 10 annotations. Overall, we use 323 attribute-object pairs (273 with colors, 42 materials, and 8 patterns).  In evaluation, every ground-truth box in OVAD is considered a possible retrieval ($\approx$14,300 samples). For CLIP, we input crops for each GT box to the image encoder and use similarity between image and text features to rank retrievals. For OVR-CNN, we compute the region embedding $f_{i}^{I}$ for each box. Then for all text \textit{t} in $\mathcal{T}$, we compute the dot product between the average word embedding $f_{j}^{t}$ of its attribute and object text tokens (\eg average($f_{red}$, $f_{car}$)) and every $f_{i}^{I}$. We report recall@$k$ (a true positive is when at least one retrieval of the correct attribute and object is within the top $k$).
    We also report precision@$k$, the proportion of correct retrievals in the top $k$. We do not directly evaluate on OVAD since the task has the different goal of detecting all attributes rather than differentiating between categories described with attributes.

    \noindent \textbf{Object attribution} For CLIP, we consider object attribution (with the VGA dataset \cite{yuksekgonul2022and}) as a relevant benchmark for image-text matching.     This task involves selecting the correct text for an image, given two choices with different order (\eg ``the crouched cat and the open door" vs. ``the open cat and the crouched door"). Note that this task is \textit{complementary} to the retrieval task, in that they both test attribute understanding, but text-region retrieval focuses more on fine-grained differentiation among \emph{plausible} attribute-object pairs (\textit{blue car} vs. \textit{red car} vs. \textit{blue truck}), while attribution focuses on intra-caption ordering where negative pairs are often \emph{implausible} (\eg ``crouched door''). 

    \noindent \textbf{Training}  Full-scale comparison to \cite{zareian2021open} uses 8 Quadro RTX 5000 GPUs and settings from \cite{zareian2021open}. For other OVR-CNN results, we pretrain using 4 NVIDIA GeForce GTX 1080 Ti with memory 11 GB. Pretraining uses 80k iter., batch size  (BS) 16, and learning rate (LR) 0.01 that scales down 10x after 40k/70k steps. For COCO finetuning, we use 4 GPUs, 
         75k iter., BS 8, and LR 0.005 that scales down after 30k/60k steps.
         CLIP finetuning is performed using OpenCLIP \cite{ilharco_gabriel_2021_5143773}, for 5 epochs using BS 64 and LR 1e-6 on 1 Quadro RTX 5000. CLIP's image encoder is ViT-B/32.

\section{Experimental results and analysis} 
    \label{sec:results}

    In Section \ref{insights}, we analyze the attribute sensitivity of VL alignment. For OVR-CNN region-word grounding, we test removing context in the captions used for pretraining detection (Fig. \ref{remove_fig}) and perturbing captions in unsupervised phrase grounding (Fig. \ref{attribute_pg_fig2}). For CLIP image-text alignment, we test perturbing the text prompts for classification via description (Fig. \ref{classviadesc_fig}). In Section \ref{evaluate}, we further evaluate how attribute context sensitivity impacts practical downstream tasks. We evaluate the impact of attribute sensitivity on an \textit{object-focused} task, in particular open-vocabulary detection with OVR-CNN (Table \ref{negative_strategies}/\ref{soa}). We also evaluate models on two \textit{fine-grained} tasks that require attribute knowledge, namely, text-region retrieval and object attribution (Table \ref{clipretrieval}/\ref{retrieval}).

    \subsection{Gauging the role of attribute context}
        \label{insights}
        \begin{figure}[t]
    \centering
    \includegraphics[scale=0.24]{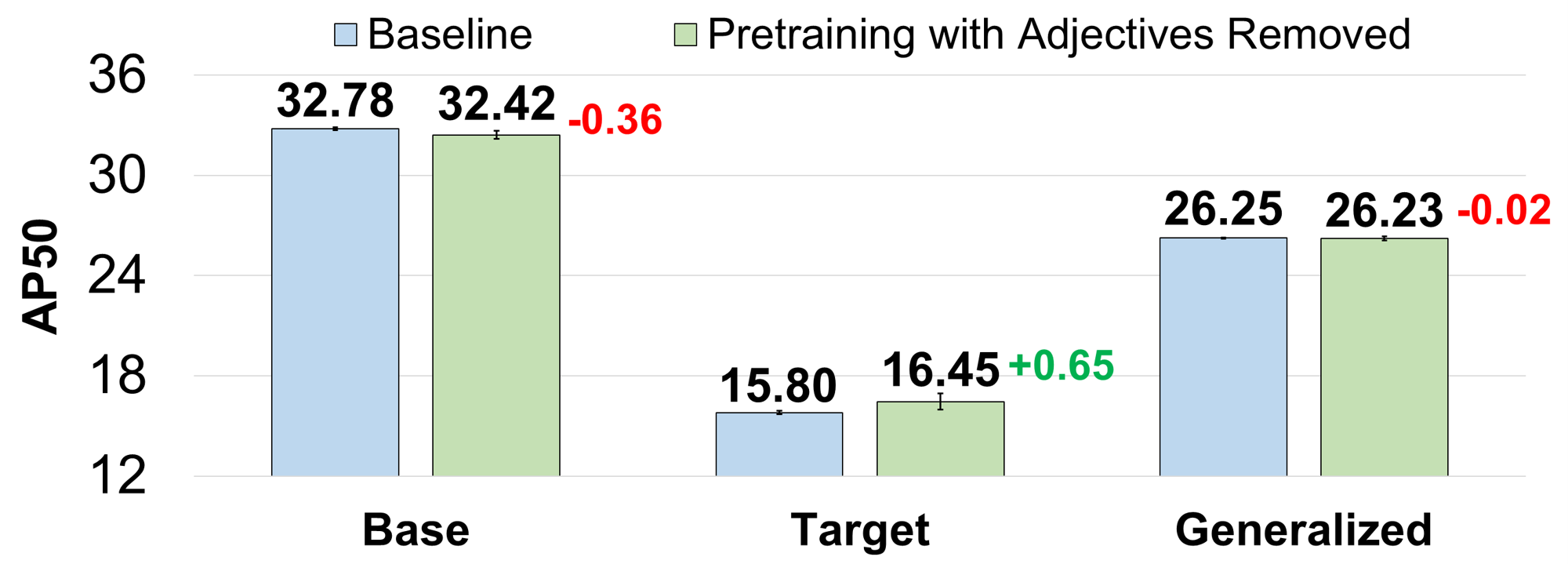}
    \vspace{-2mm}
    \caption{\textbf{Effects from removing adjectives in OVR-CNN pretraining on COCO open-vocabulary detection}. Across settings, the maximum drop from removing adjectives from training is only -0.36 AP$_{50}$. These results indicate that attribute context has limited benefit in detection. Error bars show std. error (3 trials).}
         
    \label{remove_fig}
\end{figure} 

        \begin{figure}[t]
    \centering
    \includegraphics[scale=0.245]{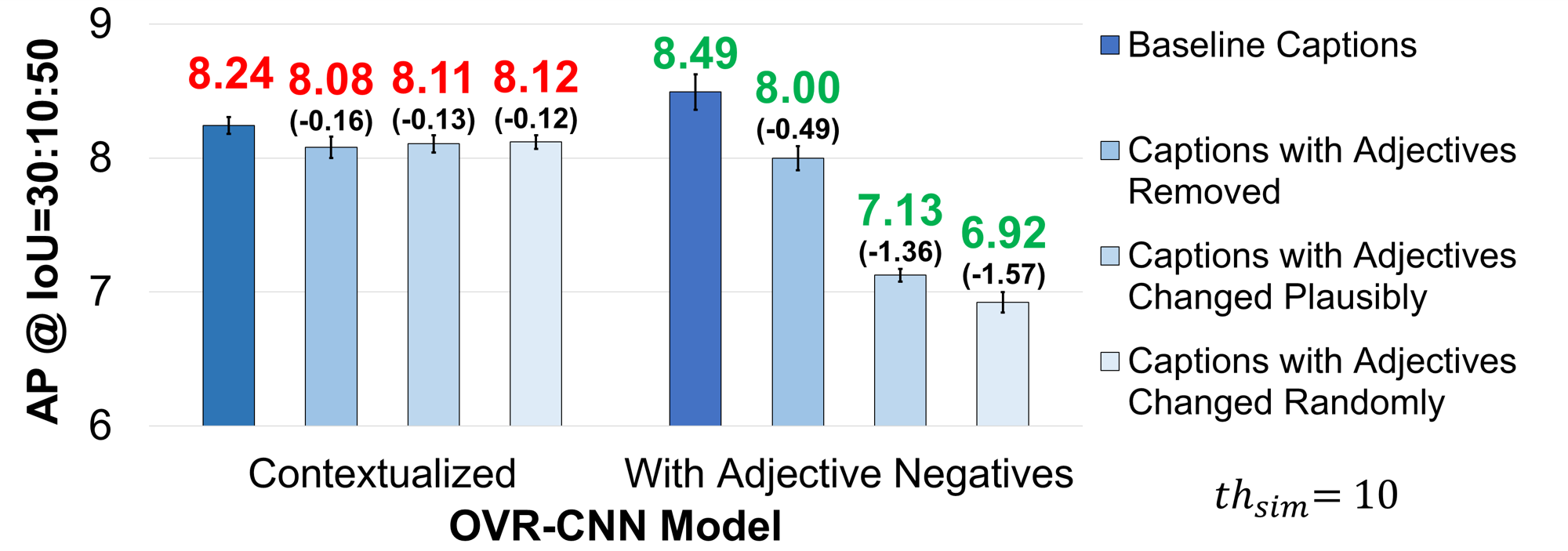}
    \vspace{-6mm}
    \caption{\textbf{Measuring attribute sensitivity in contextualized object grounding}. We find limited sensitivity to attribute meaning in default contextualized grounding, but enhanced sensitivity with (plausible) adjective negatives added. This observation is supported by  
    AP differences \textcolor{black}{(in black)} with incorrect adjectives used for the adjective negative vs. default contextualized models. The drops are discernibly larger with adjective negatives: \textcolor{Greenish}{-1.36\% } vs. \textcolor{red}{$<$0.2\% }from baseline captions to changing plausibly and \textcolor{Greenish}{-1.57\% } vs. \textcolor{red}{$<$0.2\% } from baseline captions to changing randomly. Values are avgs. over 3 training runs. Bars show std. error.}

    \label{attribute_pg_fig2}
\end{figure}  


         \setlength{\tabcolsep}{3.0pt}
\begin{table*}[t]
    \begin{center}
    \begin{tabular}{c|c|c|c|cg|cg|cgcgcg}
    
    \hline
    
    \scriptsize Adjective & \scriptsize Noun & \scriptsize Grounding & \scriptsize LE/PL  & \multicolumn{2}{c|}{\scriptsize Base-Only} & \multicolumn{2}{c|}{\scriptsize Target-Only} & \multicolumn{6}{c}{\scriptsize Generalized} \\
       \scriptsize Negative & \scriptsize Negative & \scriptsize Type & \scriptsize Trained  & \scriptsize AP$_{50}$ &  \multicolumn{1}{c|}{\scriptsize $\Delta$}  & \scriptsize \scriptsize AP$_{50}$ & \multicolumn{1}{c|}{\scriptsize $\Delta$}  & \scriptsize All AP$_{50}$ & \multicolumn{1}{c}{\scriptsize $\Delta$}  & \scriptsize Base AP$_{50}$ & \multicolumn{1}{c}{\scriptsize $\Delta$} & \scriptsize Target \scriptsize AP$_{50}$ & \multicolumn{1}{c}{\scriptsize $\Delta$}  \\
    \hline 
    \rowcolor{LLGray}
     - & - & \scriptsize Context-Free & - &  \small 32.8  \scriptsize  $\pm$ 0.08 & \scriptsize $-$ & \small 15.8  \scriptsize  $\pm$ 0.11 & \scriptsize $-$ & \small 26.3  \scriptsize  $\pm$ 0.04 & \scriptsize $-$  & \small 31.4  \scriptsize  $\pm$ 0.15 & \scriptsize $-$ & \small 11.8  \scriptsize $\pm$ 0.28 & \scriptsize $-$   \\
     \hline
    \hline
        \scriptsize Plausible &  \scriptsize \checkmark & \scriptsize Contextualized & \scriptsize \checkmark  &  \small \textbf{35.8} \scriptsize $\pm$ 0.09 & \textbf{\color{Greenish}{\small +3.0}} & \small 17.7  \scriptsize $\pm$ 0.38 & \color{Greenish}{\small +1.9} & \small \textbf{28.8}  \scriptsize  $\pm$ 0.17 & \textbf{\color{Greenish}{\small +2.5}} & \small \textbf{33.9}  \scriptsize $\pm$ 0.24 & \textbf{\color{Greenish}{\small +2.5}} & \small 14.2  \scriptsize  $\pm$ 0.34 & \color{Greenish}{\small +2.4}  \\
    \hline
        \scriptsize Random &  \scriptsize \checkmark & \scriptsize Contextualized & \scriptsize \checkmark  &    \small 35.7 \scriptsize $\pm$ 0.31 & \color{Greenish}{\small +2.9} & \small 18.0  \scriptsize $\pm$ 0.25 & \color{Greenish}{\small +2.2} & \small 28.6 \scriptsize $\pm$ 0.33 & \color{Greenish}{\small +2.3} & \small 33.5 \scriptsize $\pm$ 0.30 & \color{Greenish}{\small +2.1} & \small \textbf{14.5} \scriptsize $\pm$ 0.41& \color{Greenish}{\small \textbf{+2.7}}  \\
    \hline
            - &\scriptsize \checkmark & \scriptsize Contextualized & \scriptsize \checkmark  & \small 35.3  \scriptsize $\pm$ 0.19 & \color{Greenish}{\small +2.5} & \small 17.8  \scriptsize  $\pm$ 0.18 & \color{Greenish}{\small +2.0} & \small 28.3  \scriptsize  $\pm$ 0.12 & \color{Greenish}{\small +2.0} & \small 33.3  \scriptsize $\pm$ 0.13 & \color{Greenish}{\small +1.9} & \small 14.2  \scriptsize $\pm$ 0.13 & \color{Greenish}{\small +2.4}  \\
    \hline
            - & - &\scriptsize Contextualized &\scriptsize \checkmark & \small 35.2   \scriptsize $\pm$ 0.13  & \color{Greenish}{\small +2.4} & \small 16.7   \scriptsize $\pm$ 0.26 & \color{Greenish}{\small +0.9} & \small 28.3  \scriptsize $\pm$ 0.20 & \color{Greenish}{\small +2.0}  & \small 33.6  \scriptsize  $\pm$ 0.16 & \color{Greenish}{\small +2.2} & \small 13.1  \scriptsize  $\pm$ 0.30 & \color{Greenish}{\small +1.3} \\
     \hline
        - & - & \scriptsize Contextualized & -  & \small 31.8  \scriptsize $\pm$ 0.14  & \color{red}{\small -1.0} & \small 10.5 \scriptsize  $\pm$ 0.28 & \color{red}{\small -5.3} & \small 22.7 \scriptsize  $\pm$ 0.83 & \color{red}{\small -3.6}  & \small 28.0  \scriptsize  $\pm$ 0.98 & \color{red}{\small -3.4} & \small 7.5 \scriptsize  $\pm$ 0.47 & \color{red}{\small -4.3}\\
    \hline
    \hline
        \scriptsize Plausible & \scriptsize \checkmark & \scriptsize Context-Free &\scriptsize \checkmark  & \small 34.1  \scriptsize $\pm$ 0.21  & \color{Greenish}{\small +1.3} & \small \textbf{19.3} \scriptsize  $\pm$ 0.29 & \textbf{\color{Greenish}{\small +3.5}} & \small 28.4  \scriptsize  $\pm$ 0.17 & \color{Greenish}{\small +2.1}  & \small 33.4  \scriptsize  $\pm$ 0.19 & \color{Greenish}{\small +2.0} & \small 14.3 \scriptsize  $\pm$ 0.38 & \color{Greenish}{\small +2.5}\\
    \hline
          - & - & \scriptsize Context-Free &\scriptsize \checkmark  & \small 34.1  \scriptsize $\pm$ 0.01  & \color{Greenish}{\small +1.3} & \small 19.1 \scriptsize  $\pm$ 0.72 & \color{Greenish}{\small +3.3} & \small 28.3  \scriptsize  $\pm$ 0.27 & \color{Greenish}{\small +2.0}  & \small 33.2  \scriptsize  $\pm$ 0.12 & \color{Greenish}{\small +1.8} & \small 14.4 \scriptsize  $\pm$ 0.70 & \color{Greenish}{\small +2.6}\\
    \hline
    \end{tabular}
    \end{center}
    \vspace{-5mm}
    \caption{\textbf{Adapting OVR-CNN \cite{zareian2021open} with attribute context enhancement strategies (Sec. \ref{rwgroundcase}/\ref{enhance}): adjective/noun negative caption sampling, contextualized grounding, language encoder/projection layer training (LE/PL)}, AP$_{50}$ mean over 3 trials $\pm$ std error, $\Delta$=change vs. default OVR-CNN \cite{zareian2021open} (top row). Using adjective negatives \textit{with} contextualization yields base/generalized AP$_{50}$ increases, and top base/generalized AP$_{50}$ overall, as the model is able to take into account attribute meaning in object embeddings.}


    \label{negative_strategies}
\end{table*}

        \noindent \textbf{Attribute context has limited impact in region-word pretraining for object detection.} We first examine the role of attributes through \textit{removing all ``amod" from captions} during VL pretraining with OVR-CNN. Open-vocabulary detection results for baseline OVR-CNN \cite{zareian2021open} are shown in Fig. \ref{remove_fig}. Note that the max. drop from training with to without adjectives is -0.36 AP$_{50}$ (base), and there are not discernible drops in target/generalized settings. \textit{These results point to attribute context being wasted and not helpful when learning object grounding}, and thus serve as inspiration for our investigation of ways to boost use of attribute context.

        \noindent \textbf{Contextualizing object grounding does not result in embeddings with high sensitivity to attribute meaning.} As outlined in Sec. \ref{rwgroundcase}, we contextualize grounding in OVR-CNN as one strategy to integrate attribute context. Then through unsupervised phrase grounding, we gauge sensitivity to attribute meaning and analyze whether the attributes contextualizing an object noun (\eg ``a \textit{red} \underline{car}") impact performance. Fig. \ref{attribute_pg_fig2} shows AP@IoU=30:10:50 for (1) OVR-CNN with contextualization and (2) OVR-CNN with contextualization \textit{and} plausible adjective/noun negatives, on the four region-word grounding scenarios of interest (baseline grounding, removing adjectives, changing adjectives plausibly, and changing adjectives randomly). On the left of Fig. \ref{attribute_pg_fig2}, we find that with default contextualization, changing adjectives plausibly/randomly yields similar AP to using baseline captions or captions with removed adjectives (a max. difference of 0.13 AP@IoU=30:10:50). \textit{These observations are counterintuitive}, as embeddings can be contextualized by incorrect adjectives, yet ground similarly to when there are correct adjectives. We posit that the model may be sensitive to caption structure, where object embeddings with different adjectives are close together, and the model does not have an incentive to differentiate them. Such lack of sensitivity to attribute meaning motivates our exploration of \textit{adjective negatives}; we show the effects on the right in Fig. \ref{attribute_pg_fig2}. Contextualization aptly becomes less aligned with incorrect adjectives, 
        reaching notable drops when changing plausible/randomly with respect to the baseline
        (-1.36\%/-1.57\% respectively). In Sec. \ref{evaluate}, we show the importance of sensitivity in detection and retrieval.

        \begin{figure}[t]
    \centering
    \includegraphics[scale=0.235]{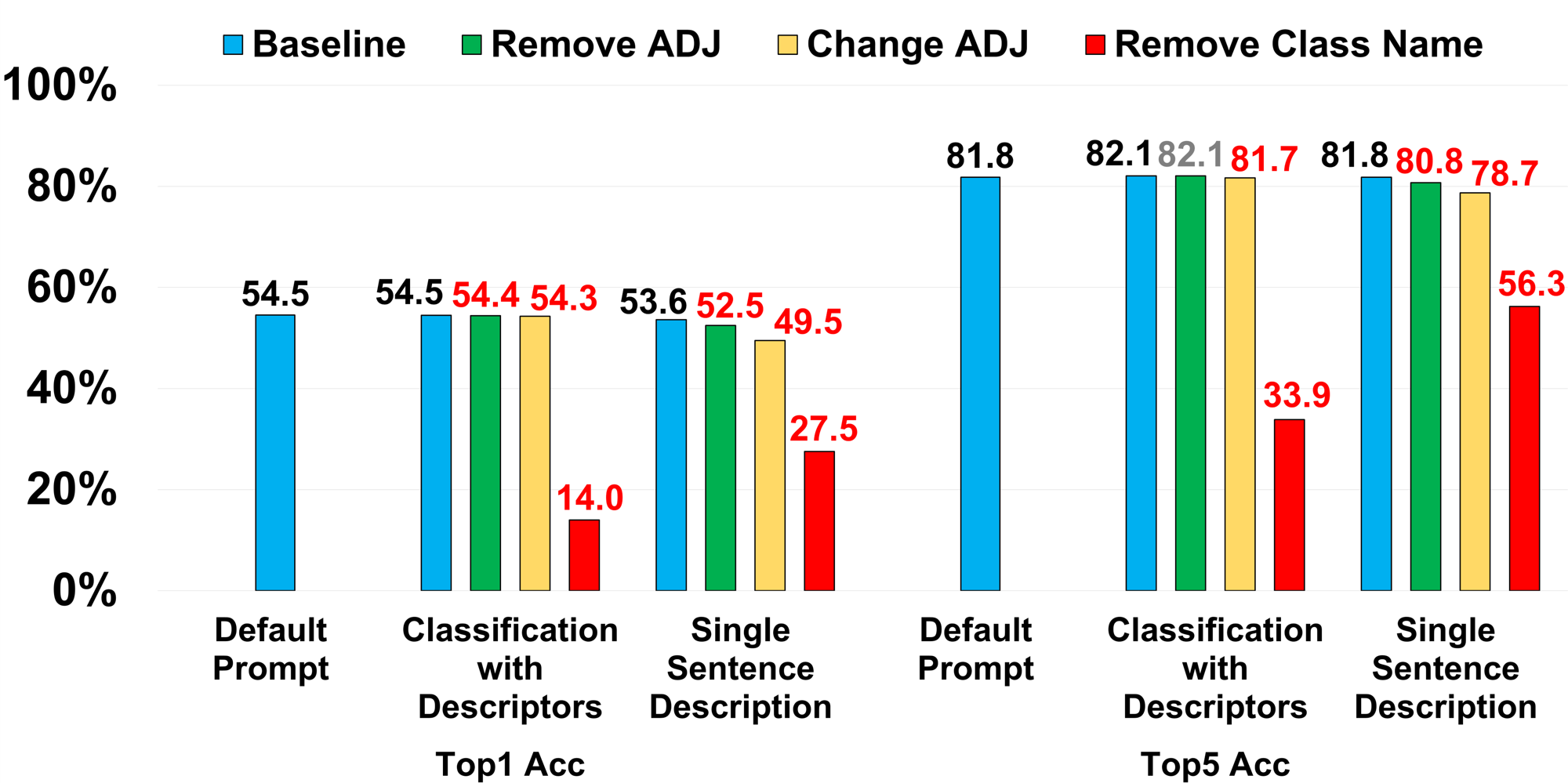}
    \vspace{-2mm}
    \caption{\textbf{Perturbing attributes and object names in CLIP descriptions used for ImageNetV2 classification.} Removing and changing adjectives have small effects on accuracy. When classes are described without object names, accuracy significantly drops.
     }
    \label{classviadesc_fig}
\end{figure}

        \noindent \textbf{Describing classes in terms of attributes alone is ineffective.} We measure CLIP's sensitivity to attributes with classification via description. As outlined in Sec. \ref{sens_method}, zero-shot inference is performed on ImageNetV2 using CLIP default prompting, LLM-based sets of object feature descriptions \cite{menon2022visual}, and LLM-based single-sentence descriptions of objects \cite{pratt2022does}. In Fig.~\ref{classviadesc_fig}, we show the results of removing/changing adjectives and removing class names in terms of top1/5 accuracy. Removing/changing adjectives results in \emph{insignificant drops} vs. the baseline with \cite{menon2022visual}, and slightly bigger drops with the \cite{pratt2022does}-like method (-4.1\% drop baseline to changing),  
        potentially as a result of more adjective-dense descriptions (supported in the supp.).
        However, \textit{removing class names} results in close to \textit{ten times more substantial} drops (max -40.5\% top1 accuracy). These results bring into question the model's ability to leverage attribute descriptions since \emph{class names drive performance}. Such results also limit the appeal of using attribute descriptions for new/custom objects with names not in the pretraining set.

    \subsection{Evaluating context enhancement strategies}
        \label{evaluate}
           
        \noindent \textbf{Enhancing context sensitivity helps open-vocabulary detection in base and generalized settings.} We evaluate OVR-CNN with four strategies to boost attribute context in region-word pretraining: (1) contextualized grounding, (2) adjective negative sampling (plausible/random), (3) noun negative sampling (random), and (4) language encoder/projection layer training, with results shown at an experimental scale in Table \ref{negative_strategies}. Compared to the baseline \cite{zareian2021open}, combining all strategies, in both plausible and random adjective negative cases, provides the largest gains in base-only and generalized (all-class) settings (\eg +3.0 and +2.5 AP$_{50}$ respectively with plausible). In Table \ref{soa}, we also present a proof-of-concept showing that enhancing attribute context improves the results reported in \cite{zareian2021open} in 4/5 settings (+0.9-1.0 AP$_{50}$ in base-only and all generalized settings). \textit{Such results highlight value in better using context, especially attributes, when learning grounding for detection}.
         \setlength{\tabcolsep}{3pt}
\begin{table}[t]
    \begin{center}
    \begin{tabular}{c|c|c|c|c|c}
    \hline
    
    \small Method & \small Base & \small Target & \multicolumn{3}{c}{\small Generalized}     \\
    & &     & \multicolumn{1}{c}{\small Base} & \multicolumn{1}{c}{\small Target} & \multicolumn{1}{c}{\small All}  \\
    \hline
    \hline

    
    \rowcolor{LLGray}
        \small \textbf{OVR-CNN} \cite{zareian2021open} & \small 46.8 & \small \textbf{27.5} & \small 46.0 & \small \small 22.8 & \small 39.9  \\
    \hline
        \small + Context Enhancement & \textbf{\small 47.7} & \small 26.5 & \textbf{\small 46.9} & \textbf{\small 23.8} & \textbf{\small 40.8}  \\
    \hline
    \end{tabular}
    \end{center}
    \vspace{-5mm}
    \caption{\textbf{OVR-CNN at full scale with various context enhancement strategies} (plausible adjective/noun negatives, contextualized grounding, language encoder/projection layer training), compared to baseline reported in \cite{zareian2021open}.  AP$_{50}$ reported on COCO.}
    \label{soa}
\end{table}

        Breaking down Table \ref{negative_strategies}, a key observation is that plausible/random adjective negatives, when used with contextualized grounding, result in (comparable) base and generalized gains over all other baselines (+0.5 and +0.4 AP$_{50}$ with plausible). These results can be ascribed to increased attention to attribute meaning that is obtainable with contextualized grounding, but not with context-free grounding since embeddings do not vary with context. \textit{There is marked benefit to learning to ground objects with attribute signals for detection.} Still, there is a tradeoff between contextualized and context-free grounding. Contextualized models result in top AP$_{50}$ in base-only and all generalized settings, but context-free results in top AP$_{50}$ in target-only. These results can be attributed to using contextualized embeddings and \textit{not} adjective negatives, since all contextualized methods obtain worse target performance than the best context-free method. We reason that the drop is due to the need for a prompt: we use a simple ``A/an $<$objName$>$." (see Sec. \ref{rwgroundcase}), but this prompt may be suboptimal to represent the large variance of contextualized embeddings for an object. Training with box annotations in base may allow visual embeddings to adjust to prompts, explaining base gains, but with no target training, adjustment cannot occur. 
        We surmise that recent work in context optimization \cite{du2022learning} can overcome this challenge. The noun negatives notably improve target-only vs. contextualized (+1.1 AP$_{50}$), showing that differentiating nouns in the same context may also help.
        
        We further inspect the plausible case by comparing class-by-class results using models in row 2/4 of Table \ref{negative_strategies}. Notably, the classes with top AP$_{50}$ gains are \textit{oven} (+4.6), \textit{bear} (+4.3), \textit{horse} (+3.6), and \textit{frisbee} (+3.4). Upon inspection of the corpus, these are commonly described in captions with visually distinctive adjectives  that may help grounding such as colors (\eg ``yellow frisbee"). Overall, we observe that 32/48 classes improve in AP$_{50}$ with adjective negatives.

        \noindent \textbf{Adjective negatives increase CLIP's fine-grained utility in multiple tasks.} We use text-region retrieval and attribution as fine-grained tasks to evaluate attribute-object understanding. Table \ref{clipretrieval} shows these results comparing strategies for finetuning CLIP on COCO: (1) choosing a random negative caption, (2) order-perturbing adjectives/nouns \cite{yuksekgonul2022and}, (3) random adjective sampling, and (4)  plausible adjective sampling. On retrieval,
        random adjective sampling is generally most effective across values of $k$, plausible is second, and both strategies outperform a random caption baseline and the order-perturbing captions of \cite{yuksekgonul2022and}. \textit{The fine-grained differentiation needed for retrieval is aided best by adjective negatives}. On the attribution task, the order-perturbing negatives perform best, which makes sense given that attribution involves determining the correct order of adjectives and nouns. It is notable that adjective negatives improve on this task \textit{and} retrieval vs. a random caption baseline, \emph{unlike the order-perturbing captions}. This shows adjective negatives achieve more generalizable attribute-object understanding across tasks. Adjective negatives similarly improve in retrieval for OVR-CNN (Table \ref{retrieval}). Plausible and random adjective sampling are more competitive in this scenario, though random sampling has highest R@1/P@1 and plausible sampling P/R@5/10. We surmise that random adjective sampling may solidify easier retrievals 
        by comparing to a wide array of adjectives, while plausible sampling may help the model differentiate between tougher cases as plausible adjectives serve as more realistic, \textit{harder} negatives.
        

         \setlength{\tabcolsep}{1.4pt}
\begin{table}
    \begin{center}
    \begin{tabular}{c|ccc|ccc||c}
    \hline 
    \small Method & \small R@1 & \small R@5 & \small R@10  & \small P@1 & \small P@5 & \small P@10 & \small VGA
    \\
    \hline
    \hline
        \scriptsize Default CLIP & \small 48.92 & \small 82.97 & \small 90.40 & \small 48.92 & \small 42.66 & \small 37.62 & \small 62.82  \\
    \hline
    \hline
        \scriptsize Random Neg. & \small 57.59 & \small \underline{87.62} & \small \textbf{94.12} & \small 57.59 & \small \underline{50.96} & \small 44.37 & \small 64.64 \\
    \hline\hline
        \scriptsize Order-Based Neg. \cite{yuksekgonul2022and}  & \small 56.97 & \small 85.76 & \small 92.88 & \small 56.97 & \small 48.73 & \small 42.79 & \color{Greenish}{\small \textbf{73.87}} \\
    \hline
    \hline
        \scriptsize Plausible Adj. Neg.  & \color{Greenish}{\small \underline{58.82}} & \small 86.69 & \small \underline{93.81} & \small \color{Greenish}{\underline{58.82}} & \small \underline{50.96} & \small \color{Greenish}{\underline{44.77}} & \color{Greenish}{\small \underline{67.94}} \\
    \hline
        \scriptsize Random Adj. Neg.  & \small \color{Greenish}{\textbf{60.06}} & \color{Greenish}{\small \textbf{88.24}} & \small 92.26 & \small \color{Greenish}{\textbf{60.06}} & \color{Greenish}{\small \textbf{51.76}} & \color{Greenish}{\small \textbf{44.98}} & \color{Greenish}{\small 67.93} \\
    \hline

    \end{tabular}
    \end{center}
    \vspace{-5mm}
    \caption{\textbf{Fine-grained utility of CLIP finetuned with negative sampling strategies, on T2R retrieval and Visual Genome Attribution (VGA) \cite{yuksekgonul2022and}.} Recall/precision@$k$=1,5,10 are reported for T2R retrieval and accuracy for VGA. Best=\textbf{bold}, second=\underline{underlined}, results $>$ random baseline (row 2) in \color{Greenish}{green}\color{black}. Note that adjective sampling offers improvements across \textit{both} attribute tasks, while order only helps on the order-based VGA task.}
    \label{clipretrieval}
\end{table}

         \setlength{\tabcolsep}{1.75pt}
\begin{table}
    \begin{center}
    \begin{tabular}{c|ccc|ccc}
    \hline 
    \small Method & \small R@1 & \small R@5 & \small R@10  & \small P@1 & \small P@5 & \multicolumn{1}{c}{\small P@10} 
    \\
    \hline
    \hline
        \scriptsize Contextualized Baseline  & \small 13.21 & \small 37.36 & \small 52.01 & \small 13.21 & \small 12.84 & \small 11.75  \\
    \hline
    \hline
        \scriptsize Plausible Adjective Negative  & \small \underline{16.82} & \small \textbf{39.83} & \small \textbf{53.35} & \small \underline{16.82} & \small \textbf{14.14} & \small \textbf{12.63} \\
    \hline
        \scriptsize Random Adjective Negative  & \small \textbf{17.44} & \small \underline{39.42} & \small \underline{52.53} & \small \textbf{17.44} & \small \underline{13.70} & \small \underline{12.32} \\
    
    \hline
    \end{tabular}
    \end{center}
    \vspace{-5mm}
    \caption{\textbf{Fine-grained utility of OVR-CNN, pretrained with adjective negatives, in text-region retrieval of attribute-object concepts.} 
    Recall/precision@$k$=1,5,10 are reported over 3 trials. }
    \label{retrieval}
\end{table}




\section{Conclusion}

    \label{sec:conclusion}

    We answer these questions (Sec. \ref{sec:intro}):
    (1) Attribute context can show limited impact in region-word pretraining for detection. (2) Grounding objects to contextualized word embeddings increases attribute consideration only to a limited degree.
    (3) Describing CLIP's classes by only their attributes results in poor accuracy. 
    Also, models struggle at fine-grained retrieval.
    (4) Adjective-based negative caption sampling is promising to increase model sensitivity to attribute meaning and especially boosts fine-grained retrieval.
    (5) Plausible and random adjective sampling are competitive in detection/retrieval following OVR-CNN grounding; with CLIP, random sampling has higher retrieval gains.

    \noindent \textbf{Acknowledgements:} This work was supported by a National Science Foundation Grant No. 2006885.





{\small\bibliographystyle{ieee_fullname}\bibliography{egbib}}

\clearpage

\thispagestyle{plain}

\section{Supplementary Material}
The supplementary material contains two sections. Section \ref{datagen_supp} outlines how we generate data for training and evaluation. Section \ref{pg_supp} shows further experiments for phrase grounding analysis of OVR-CNN.

\subsection{Data Generation}
\label{datagen_supp}

\subsubsection{Classification via description}

We present details regarding data generation for measurement of CLIP's attribute sensitivity, particularly for use in the classification by description task on ImageNet-v2 \cite{recht2019imagenet}. Overall, three ``styles" of CLIP prompts are used in inference: (1) CLIP's default ``a photo of" prompts, (2) LLM-based \textit{multiple descriptor} prompts, and (3) LLM-based \textit{single sentence} prompts.

To produce (1), all class names from ImageNet-v2 are filled into the following template and processed by CLIP's text encoder to produce classifier weights:

\begin{quote}
``a photo of a $<$category$>$."
\end{quote}

For instance, CLIP's classifier would contain the text encodings of  ``a photo of a petri dish", ``a photo of a basketball", etc. (all 1,000 classes).

For (2), we use the methodology of Menon and Vondrick \cite{menon2022visual} to produce descriptions with attribute context. For every class in ImageNet-v2, we 
prompt GPT-3 (\textit{davinci-002}, max token length 100, temperature 0.7) with a \textit{multiple descriptor} prompt template: 

\begin{quote}
   Q: What are useful features for distinguishing a lemur in a photo? \\
    A: There are several useful visual features to tell there is a lemur in a photo: \\
    - four-limbed primate \\
    - black, grey, white, brown, or red-brown \\
    - wet and hairless nose with curved nostrils \\
    - long tail \\
    - large eyes \\
    - furry bodies \\
    - clawed hands and feet \\
    Q: What are useful features for distinguishing $<$category$>$ in a photo? \\
    A: There are several useful visual features to tell there is/are $<$category$>$ in a photo:
\end{quote}

Output descriptors are returned in the format of the lemur example. We further process these outputs with the following CLIP prompt template:

\begin{quote}
    $<$category$>$ which (is/has/etc) $<$descriptor$>$.
\end{quote}

There are thus multiple prompts for each class. There would be seven prompts for the lemur example, for instance. These would appear as:

\begin{quote}
    - a lemur which (is/has/etc) a four-limbed primate. \\
    - a lemur which (is/has/etc) black, grey, white, brown, or red-brown. \\
    - ... 
\end{quote}

The average CLIP similarity between the image of interest and each descriptor (Eq. 7 in main paper) is used as the score for each class in classification.

For (3), we use a \textit{single-sentence} description-based prompt template for GPT-3, in particular the one from \cite{pratt2022does}: 

\begin{quote}
    Q: What does a lorikeet look like? Describe with one sentence. \\
    A: A lorikeet is a small to medium-sized parrot with a brightly colored plumage. \\
    Q: What does $<$category$>$ look like? Describe with one sentence. \\
    A: 
\end{quote}

We directly use each class's result from GPT-3 as a respective CLIP prompt. For instance, the lorikeet's CLIP prompt would be:

\begin{quote}
    A lorikeet is a small to medium-sized parrot with a brightly colored plumage.
\end{quote}

The lemur's CLIP prompt could be: 

\begin{quote}
    A lemur is a small, four-limbed primate with large eyes, a long tail, and a slender body. 
\end{quote}

After generating the prompts for (1)-(3), we remove/change adjectives detected with
 spaCy \cite{Honnibal_spaCy_Industrial-strength_Natural_2020} (v3.5.3) in all class prompts for inference. For class name removal, we replace \textit{all} class names with ``a/an object" when creating the CLIP classifier. As an example, consider possible prompts used to create CLIP's class embeddings with (3): 

 \begin{quote}
    (P1) A baseball is a round, stitched ball made of leather or synthetic materials, typically white with red stitching. \\
    (P2) A hockey puck is a flat, disk-shaped object made of hard rubber, often black in color, used in the sport of ice hockey. \\
    ...\\
    (P1000) A basketball is a round, inflatable ball with a synthetic or leather cover, black lines, and typically orange in color.
 \end{quote}

 Removed class names would change the classifier to: 
 \begin{quote}
    (P1) An object which is a round, stitched ball made of leather or synthetic materials, typically white with red stitching. \\
    (P2) An object which is a flat, disk-shaped object made of hard rubber, often black in color, used in the sport of ice hockey. \\
    ...\\
    (P1000) An object which is a round, inflatable ball with a synthetic or leather cover, black lines, and typically orange in color.
 \end{quote}

In Table \ref{tab:cvd}, we provide key statistics regarding the descriptors generated for this analysis.

\begin{table}[h]
\label{cvd}
    \begin{center}
        \renewcommand{\arraystretch}{1}
        \begin{tabular}{|c|c|c|}
            \hline
                \small \textbf{Statistic} & \small \textbf{CWD \cite{menon2022visual}} & \small \textbf{SS \cite{pratt2022does}} \\
                 
                \hline\hline
                \small Avg. \# Descriptions Per Class  & \small 5.29 & \small 1  \\
                \hline
                \small Avg. Desc. Length (spaCy tokens)  & \small 18.43 & \small 20.65  \\
                \hline
                 \small Total \# Adjectives Perturbed  & \small 4,831 & \small 2,392 \\
                \hline
                 \small \# Unique Adjectives  & \small 607 & \small 501  \\
                \hline
                \small \# Adjectives Per Description & \small 0.91  & \small 2.39  \\
                \hline
        \end{tabular}
    \end{center}
    \vspace{-5mm}
    \caption{Statistics for measuring attribute sensitivity in classification via description. CWD=Classification With (Multiple) Descriptors; SS=Single Sentence. Note that there are more adjectives/description in the single-sentence case, potentially explaining its increased sensitivity to ``Change ADJ" in Fig. 5 (main paper).}
    \label{tab:cvd}

\end{table}

\subsubsection{Analyzing COCO}

    Given the role of COCO \cite{chen2015microsoft, lin2014microsoft} in pretraining OVR-CNN, finetuning OVR-CNN/CLIP, and gauging the attribute sensitivity of OVR-CNN, we provide specific details regarding its usage. We describe in detail how we identify objects of interest and attribute context belonging to those objects, as well as statistics related to the data creation process.
    
\begin{figure*}[h]
    \centering
    \includegraphics[scale=0.42]{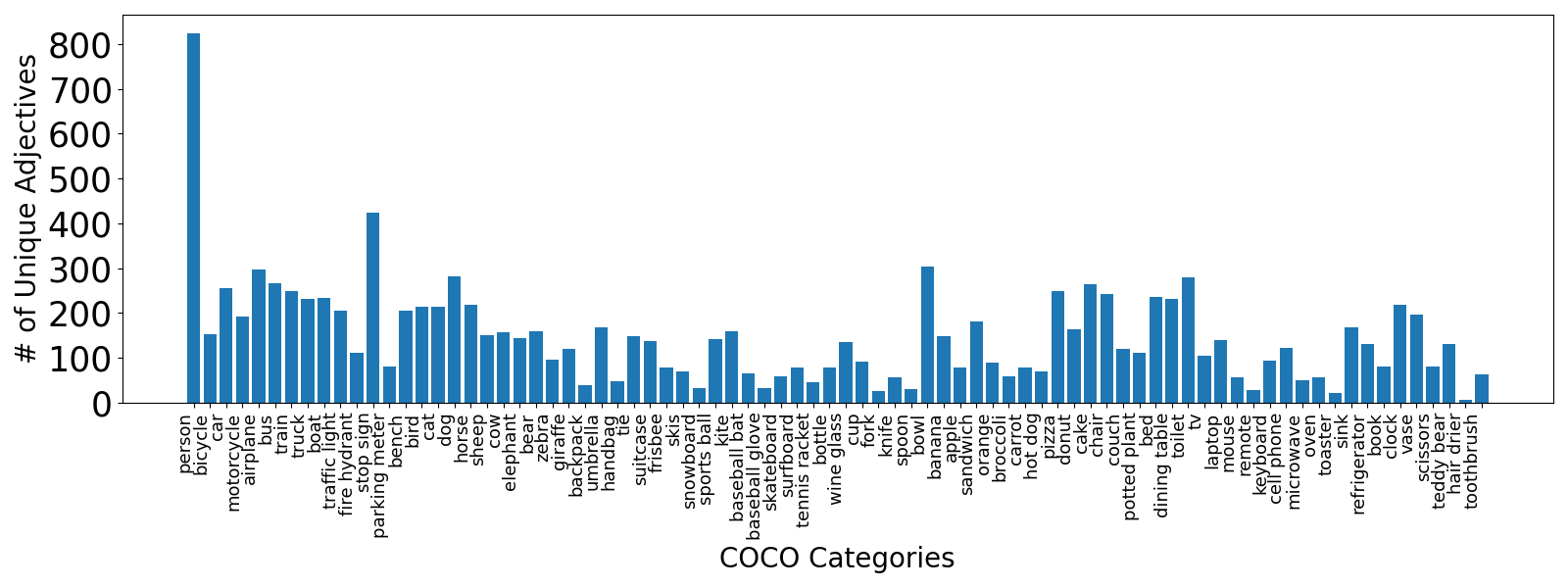}
    \caption{A histogram of unique adjectives described with objects in the COCO Captions corpus. }
    \label{coco_adj}
\end{figure*}

    For all of these tasks, we create a vocabulary $\mathcal{V}$ of terms/phrases corresponding to COCO class names (\eg ``car", ``fire hydrant", ``teddy bear"). We build $\mathcal{V}$ from the synonym list of COCO class names provided in \cite{lu2018neural}, with plural terms also added. Adjectives used for specific COCO objects are detected using the spaCy dependency parser \cite{Honnibal_spaCy_Industrial-strength_Natural_2020}. For each caption, we traverse the parse tree and mark all ``amod" with dependency on a class term in $\mathcal{V}$, taking note of which class each term belongs to. We also mark terms not explicitly detected as ``amod", but connected through coordinating conjunctions (``cc"). We use per-class lists to create the \textit{plausible} sets. Some example top occurring plausible adjectives are shown in Table \ref{tab:counts_attr}. All unique adjectives detected across all classes lie in the \textit{random} set. We show the counts of unique adjectives detected across COCO classes in Fig. \ref{coco_adj}.  These sets are used to sample negative caption adjectives in pretraining and to change plausibly/randomly in unsupervised phrase grounding.

    \begin{table}[h]
\label{cvd}
    \begin{center}
        \renewcommand{\arraystretch}{1}
        \begin{tabular}{|c|c|c|}
            \hline
                \small \textbf{COCO Class} & \small \textbf{Adjective} & \small \textbf{Count} \\
                \hline\hline 
                 \small bear & \small black & \small 860 \\
                \small & \small brown & \small 822 \\
                \small & \small polar & \small 802 \\
                \small & \small large & \small 486 \\
                \small & \small white & \small 244 \\
                \hline
                \small frisbee & \small white & \small 302 \\
                \small & \small yellow & \small 229 \\
                \small & \small red & \small 201 \\
                \small & \small blue & \small 132 \\
                \small & \small green & \small 92 \\
                \hline\small apple & \small green & \small 182 \\
                \small & \small red & \small 153 \\
                \small & \small sliced & \small 38 \\
                \small & \small yellow & \small 26 \\
                \small & \small several & \small 22 \\
                \hline
        \end{tabular}
    \end{center}
    \vspace{-5mm}
    \caption{Examples of top 5 detected adjectives and counts for COCO categories. Note the frequent use of colors and state adjectives (\eg \textit{sliced, large}). }
    \label{tab:counts_attr}

\end{table}
    \begin{table}[h]
\label{coco_stats}
    \begin{center}
        \renewcommand{\arraystretch}{1}
        \begin{tabular}{|c|c|c|}
            \hline
                \small \textbf{Statistic} & \small \textbf{COCOtrain} & \small \textbf{COCOval} \\
                 
                \hline\hline
                \small \# Total Captions  & \small 591,753 & \small 5,000  \\
                \hline
                \small \# Caps. with COCO AdjMod  & \small 153,207 & \small 1,294  \\
                \hline
                \hline
                 \small Total \# Adjectives Perturbed  & \small 191,772 & \small 1,611 \\
                \hline
                 \small \# Unique Adjectives  & \small 3,080 & \small 277  \\
                \hline
        \end{tabular}
    \end{center}
    \vspace{-5mm}
    \caption{\textbf{Statistics for modifying adjectives in COCO captions (train/val).} AdjMod = token labeled as an adjectival modifier. Many captions ($>$150k) contain one or more adjectives, which represents a significant amount of signal that models can leverage.}
    \label{coco_stats}

\end{table}
    \begin{figure*}[t]
    \centering
    \includegraphics[scale=0.225]{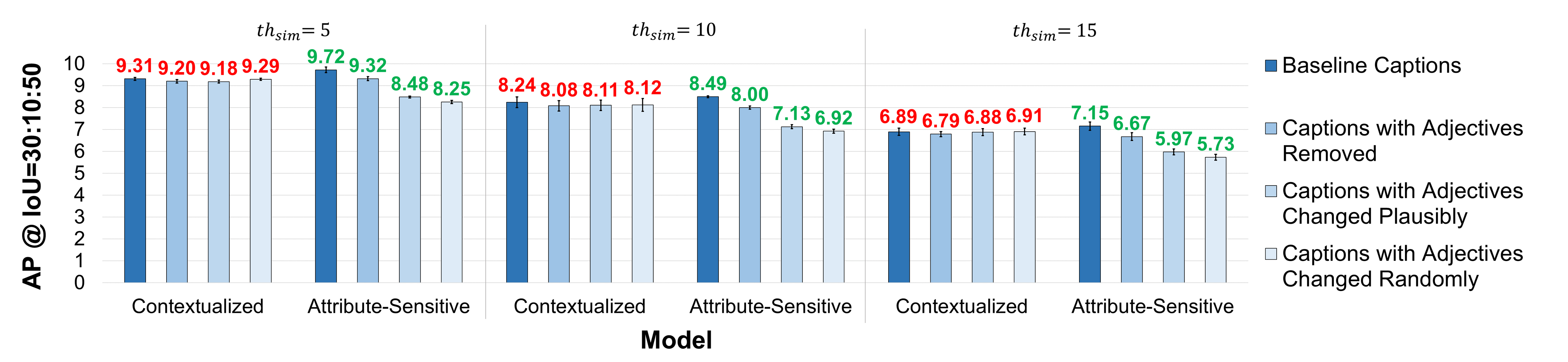}
    \vspace{-3mm}
    \caption{\textbf{Measuring attribute sensitivity in contextualized object grounding}. In an attribute-sensitive model, grounding performance should drop if an incorrect attribute is used, which occurs when changing adjectives. Through unsupervised grounding, we show that default contextualization does \textit{not} result in substantial AP@IoU=30:10:50 drops vs. the baseline with adjectives changed (\textcolor{red}{red}), illustrating a lack of sensitivity to attribute meaning. When we add adjective negatives (plausible in this case), contextualization gains enhanced sensitivity to attribute meaning, shown in the decreases from baseline to changing (\textcolor{Greenish}{green}). \textbf{Note that such trends hold over values of $th_{sim}$}. The presented values are averages over 3 pretraining trials, and error bars show standard error.}
    \label{pg_supp_thresh}
\end{figure*} 
    
    For OVR-CNN, COCO is used in our study in pretraining (2017train as an image-caption dataset), in finetuning (2017train/2017val for detection), and in unsupervised phrase grounding (2017val as an image-caption evaluation set). For CLIP, COCO is used in finetuning (2017train/2017val for image-text matching). COCO is used with both models for text-region retrieval (2017val). We provide further details of the detected adjectives in train/val in Table \ref{coco_stats}. For phrase grounding, we choose one caption for each image to use, resulting in 5,000 test cases.

\subsection{Threshold experiments: Unsupervised phrase grounding}
\label{pg_supp}

A hyperparameter for unsupervised phrase grounding is $th_{sim}$, which determines how bounding boxes are created from similarity maps. We find in practice that the attribute sensitivity trends of interest (i.e. relative AP@$t$ differences between baseline/removal and baseline/changing) hold across values of this parameter for default contextualization and a model with adjective negatives. Fig. \ref{pg_supp_thresh} shows three values we experiment with to support this finding.


\end{document}